\documentclass[runningheads]{llncs}
\usepackage[T1]{fontenc}
\usepackage{graphicx}
\usepackage{tabularx}
\usepackage{caption}
\usepackage{subcaption}
\usepackage{tabularx}
\usepackage[table,xcdraw]{xcolor}
\usepackage{array}
\usepackage{threeparttable}
\usepackage{url}
\usepackage{amsmath,amssymb}
\begin{document}
\title{Prints in the Magnetic Dust: Robust Similarity Search in Legacy Media Images Using Checksum Count Vectors}
\titlerunning{Similarity Search in Legacy Media Images Using Checksum Count Vectors}
%
\author{Maciej Grzeszczuk\inst{1,2}\orcidID{0000-0002-9840-3398} \and
Kinga Skorupska\inst{1}\orcidID{0000-0002-9005-0348} \and
Grzegorz M. Wójcik\inst{3}\orcidID{0000-0002-4678-9874}}
\authorrunning{M. Grzeszczuk et al.}
%
\institute{Polish-Japanese Academy of Information Technology, Warsaw, Poland \and
The Foundation for the History of Home Computers, Warsaw, Poland
\email{maciej.grzeszczuk@fhkd.pl}  -  
\url{https://fhkd.pl/} \and
University of Maria Curie-Skłodowska, Lublin, Poland}
\maketitle              
\vspace{-0.5cm}
\begin{abstract}
Digitizing magnetic media containing computer data is only the first step towards the preservation of early home computing era artifacts. The audio tape images must be decoded, verified, repaired if necessary, tested, and documented. If parts of this process could be effectively automated, volunteers could focus on contributing contextual and historical knowledge rather than struggling with technical tools.
We therefore propose a feature representation based on Checksum Count Vectors and evaluate its applicability to detecting duplicates and variants of recordings within a large data store. The approach was tested on a collection of decoded tape images (n=4902), achieving 58\% accuracy in detecting variants and 97\% accuracy in identifying alternative copies, for damaged recordings with up to 75\% of records missing. These results represent an important step towards fully automated pipelines for restoration, de-duplication, and semantic integration of historical digital artifacts through sequence matching, automatic repair and knowledge discovery.

\end{abstract}

\section{Introduction}

Han shot first. At least until the 1997 Special Edition of Star Wars, in which the smuggler fires in self-defense, after the bounty hunter misses him from across the table. In response to fan outrage, the scene was altered again in the 2004 release, this time showing both shots fired almost simultaneously. On DVD and streaming platforms, only the latest version of the film remains available. Thanks to the dedication of the fan community, however, initiatives such as the Harmy's Despecialised Edition\footnote{Facebook page for the project here: \url{https://www.facebook.com/despecialized/}} and Project 4K77\footnote{Web page for the project: \url{https://www.thestarwarstrilogy.com/project-4k77/}} have reconstructed the original cuts in modern quality standards, combining and restoring fragments from surviving 35mm prints and LaserDiscs. The SETI@home project, which distributed radio telescope data analysis tasks among volunteers searching for extraterrestrial signals, had more than five million participants \cite{parsons2004seti}. In Transcribe Bentham, volunteers continue to manually convert tens of thousands of scanned historical manuscripts into digital text form usable for further study \cite{moyle2010transcribe}. Harnessing human potential - with the right tools at hand - can help reconstruct artifacts of the early home computer era \cite{grzeszczuk2023preserving}. A crowd of contributors can digitize the contents of slowly decaying magnetic media with its unique programs or user data. Some of these artifacts were created in specific economic or regional contexts, such as Eastern Europe with limited flow of technology and information due to the Iron Curtain \cite{lekkas2014legalpirates,stachniak2015redclones,wasiak2014playing} and hence, ought to be preserved to stand witness to the inception of our modern computerized world \cite{grzeszczuk2023preserving}.

\noindent\textbf{Our study}. We propose a robust vector-based approach for identifying and matching fragments of decoded digital recordings, such as magnetic tapes used with 8-bit Atari computers. Using 4902 decoded tape files we then validate the applicability of the solution in two areas of interest - \textbf{(1) damaged file reconstruction} and \textbf{(2) file identification, de-duplication and documentation}.

\paragraph{Research Questions:}

\begin{itemize}
    \item\textbf{RQ1:}\label{rq1} Can we use a set of 8-bit cassette records check sums to quickly identify the file as a duplicate or a variant of a different tape file in our records?
    \item\textbf{RQ2:}\label{rq2} Will this mechanism work correctly if one or both of the compared files are corrupted or truncated, and to what extent?
    \item\textbf{RQ3:}\label{rq3} Can the dimensionality of the vector space be reduced without negative impact on performance?
\end{itemize}

\section{Related Work}

Work on cultural heritage preservation is increasingly collaborative and distributed \cite{weber2023digital}, moving from museums and cultural institutions to the hobbyist space online \cite{chng2023photogrammetry,wirtz2025savingukraine,pereda2025onlineheritage}. Such participation models enable enthusiasts to act as vernacular archivists \cite{swalwell2021,dauterive2025save}, contributing their unique know-hows to distributed preservation efforts. However, citizen science and preservation efforts require a scaffolding to stay sustainable and motivating \cite{afforce_2022}. Citizen science platforms, such as Zooniverse \cite{zooniverse_2023}, can serve this purpose, so can specific initiatives, such as the European Holocaust Research Infrastructure (EHRI). In its work of integrating archives scattered by war and deliberate destruction, it uses methods of combining digital representations of objects from different sources in order to rebuild lost historical context \cite{Erez.2020}.

Despite large participation, the growing volume of digitized and born-digital data \cite{capaccioni2025borndigital}, and its often unstructured nature \cite{sadia2025unstructured}, can be overwhelming. Text mining techniques have been successfully used for years by cultural institutions to connect and fill gaps in the networks that connect objects with each other and with their associated metadata through countless connections \cite{aletras2013similarity}. Texts are compared with the use of cosine similarity measures calculated on their vectorized form \cite{rahutomo2012cosine}. For audio files, features derived from spectrograms, such as constellations of power-frequency peaks, can be used to efficiently identify a song from a short, noisy fragment, using a conceptually similar search mechanism \cite{DBLP:conf/ismir/Wang03,six2022panako}.

The same logic of reconstruction extends to the early home computing systems, such as Atari. Atari computers store data on magnetic tape in the form of zeros and ones represented as sounds of two different frequencies\footnote{Frequency Shift Keying modulation (FSK). It uses 5237Hz tone to represent a binary "1" (also called \textit{"mark"}) and 3995Hz tone for a binary "0" (a \textit{"space"}).}. For each byte of data there are 10 bits written on the tape (a start bit of "0" and stop bit of "1" is being used for synchronization), and the file is divided into 132-byte records (with a usable payload of 128 bytes). Rolling 8-bit checksum with carry is used on packet level for error detection \cite{Crawford1982DeReAtari}. Recovering data from such obsolete formats is both a technical feat and an act of cultural reconstruction \cite{vries2016commodore,Lichnerowicz_Grzeszczuk_Skorupsk_2024}. As the signal is extremely narrow-band, non-redundant and has low entropy, music fingerprinting methods cannot be effectively used to aid. 

\section{Methods}

\subsection{Feature Definition}

We therefore needed to create a feature that enables efficient searching for files likely to be copies of the same recording, even if different fragments are missing. While a weak 8-bit checksum of a single record does not provide sufficient redundancy to reconstruct unreadable data within the record, we can use the distribution of correctly decoded ones over all 256 checksum values to form a vector representation. Each file is therefore represented as a 256-dimensional vector that can be compared with others in this feature space.

\paragraph{Checksum count vector and similarity measure.}

Let a file $F$ consist of $N$ correctly decoded records $\{r_1, r_2, \dots, r_N\}$, 
each with an 8-bit checksum value $c_i \in \{0, 1, \dots, 255\}$.

We define the \emph{checksum count vector} 
$\mathbf{v}_F \in \mathbb{R}_{\ge 0}^{256}$ as
\begin{equation}
    \mathbf{v}_F[k] = \sum_{i=1}^{N} \mathbf{1}_{\{c_i = k\}}, 
    \quad k = 0, 1, \dots, 255,
\end{equation}
where $\mathbf{1}_{\{c_i = k\}}$ denotes the indicator function, equal to~1 
if the $i$-th record has checksum~$k$, and~0 otherwise.

The similarity between two files $F$ and $G$ can then be expressed 
as the cosine similarity between their vectors:
\begin{equation}
    S(F,G) = \frac{\mathbf{v}_F \cdot \mathbf{v}_G}
    {\|\mathbf{v}_F\|\,\|\mathbf{v}_G\|}.
\end{equation}

Files with higher $S(F,G)$ values are therefore expected to represent variants 
or partial copies of the same recording. Later in the paper we will validate our hypothesis.

\subsection{Validation and Calibration}

To validate our method, we analyzed a collection of 4902 Atari 8-bit recordings, correctly decoded into the digital .CAS format\footnote{CAS format definition: \url{https://a8cas.sourceforge.net/format-cas.html}}. Files were collected from: the Atari 8-bit Software Preservation Initiative\footnote{Web site address: \url{https://www.a8preservation.com/}} (1,229 files), Atarimania\footnote{Web site address: \url{https://www.atarimania.com/}} (1,095 files), Atari On-Line\footnote{Kaz repository had been used, as more complete than the TOSEC-convention repository from the same site: \url{https://atarionline.pl/v01/index.php?ct=kazip}} (2,101 files), and the TOSEC Database\footnote{Web site of the project: \url{https://www.tosecdev.org/}} (477 files) in the complete scope available.

\begin{figure}[htbp!]
\centering
\vspace{-10pt}
\includegraphics[width=\textwidth]{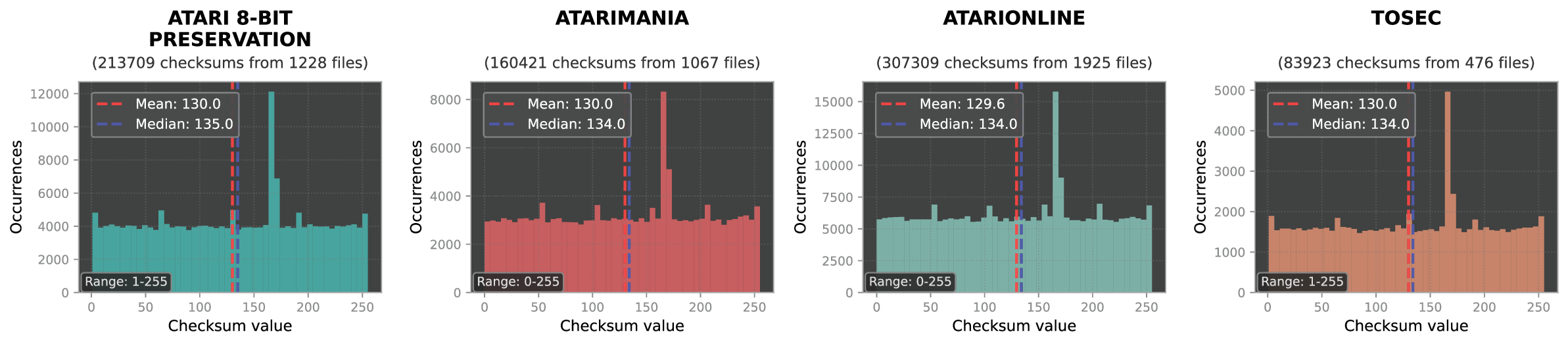}
\caption{Distribution of record checksum values for all the CAS files in the repository, by source. Visible spikes represent checksum values of 167 and 169.}
\vspace{-15pt}
\label{dystrybucja-czeksum}
\end{figure}

The .CAS files were then converted to textual .HEX form\footnote{HEX format definition: \url{https://a8cas.sourceforge.net/format-hex.html}} with the use of \textit{a8cas-util}\footnote{Website of the A8CAS-UTIL project: \url{http://www.arus.net.pl/FUJI/a8cas-util/}} decoder and processed in Python. We examined the distributions of checksums among valid records (see Figure \ref{dystrybucja-czeksum}) and investigated two significant peaks that stand out. For 167, the main source (86\%) were packets with all-zeros payload\footnote{All-zero records often carry background fill for uncompressed images.}, while the 169 were mainly the end-of-file (EOF) markers \cite{Crawford1982DeReAtari}.

\paragraph{Variants detection tests.} For this test we cleaned all the filenames of the TOSEC-style metadata and created a blacklist to filter out noise such as multi-side titles and common name compilations. The remaining 3320 filenames had been lower-cased, vectorized (TfidfVectorizer) and compared in pairs (5,509,540), to find 3955 with title cosine similarity of 0.8+. For those we calculated our \textit{checksum count vector} similarity in 256-dimensional space.

\begin{figure}[htbp!]
\centering
\vspace{-10pt}
\includegraphics[trim={0 0 10.9cm 0.5cm},clip,width=\textwidth]{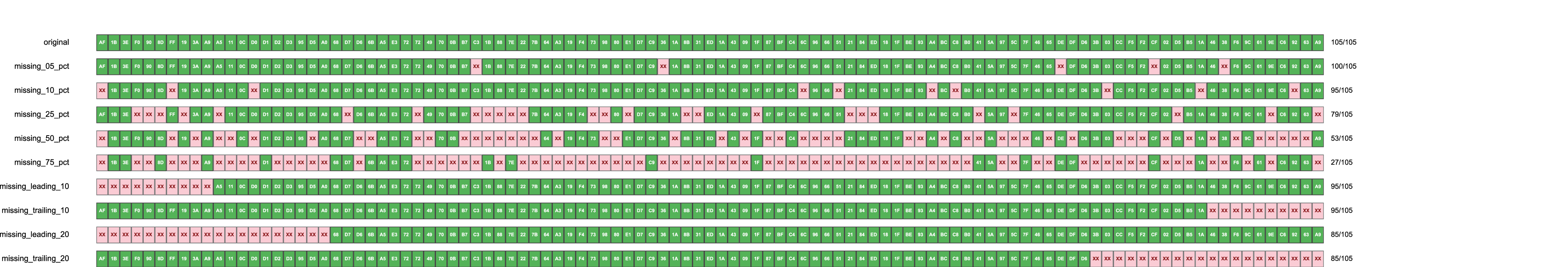}
\caption{File: \texttt{Wumpus Adventure - Expanded Version v8.7b (1981-03-19) (Sebree's Computing)(US)[CLOAD+RUN][BASIC].hex} and its 9 damaged derivatives created according to the test procedure. Missing records in pink.}
\vspace{-12pt}
\label{dziurawe-pliki}
\end{figure}

\paragraph{Damaged copy recognition.} We randomized a 10\% sample from each source (a total of 488 files) and prepared 9 variations of each file, representing different damage scenarios: 5\%, 10\%, 25\%, 50\% and 75\% random records missing, leading 10 and 20 records missing and trailing 10 and 20 records missing (see Figure \ref{dziurawe-pliki}). With that set, we performed our \textit{checksum count vector} similarity calculations in 256-, 128-, 64-, 32- and 16-dimensional space, comparing each to \textbf{(1) it's own, one grade less damaged copy} (i.e. 5\% damaged with the original file, 75\% damaged with 50\% damaged) and \textbf{(2) all the other files in the dataset, regardless of the damage}. We then repeated the tests excluding the 167 and 169 dimensions from calculations.

\section{Results}

The results showed a slight improvement in the second test pass, which was performed after excluding checksum values 167 and 169 from the vectors. Unless otherwise noted, all results reported below refer to this adjusted second pass.

\begin{figure}[htbp!]
\centering
\vspace{-12pt}
\includegraphics[width=\textwidth]{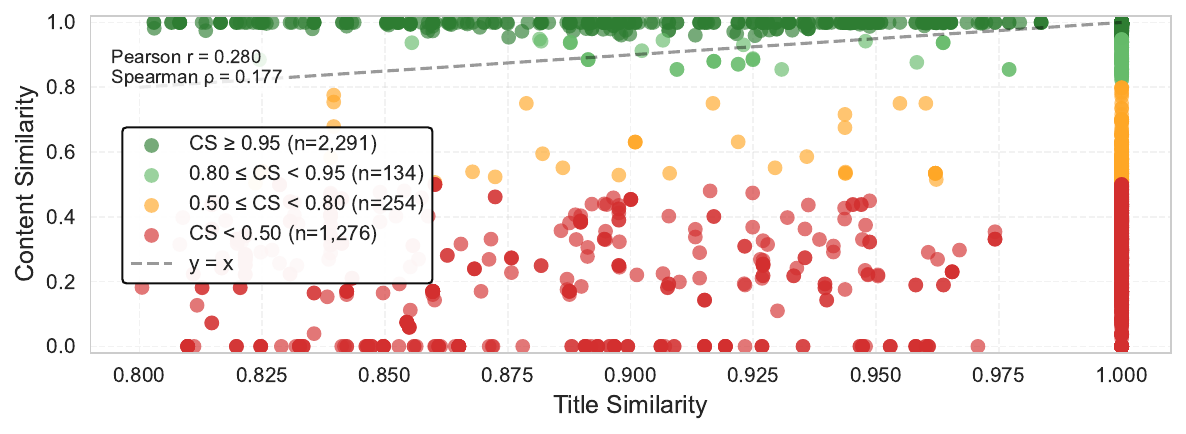}
\caption{\textbf{Correlation between title similarity and content similarity for file pairs.}
Each point represents a pair of files with title similarity $\geq 0.8$ (cosine similarity on title strings). Content similarity (CS) is calculated using 256-dimensional vectors. Colors indicate content similarity ranges: dark green (CS $\geq 0.95$),
light green ($0.80 \leq$ CS $< 0.95$), orange ($0.50 \leq$ CS $< 0.80$), and red (CS $< 0.50$). The dashed line represents perfect correlation ($y=x$). Pearson correlation coefficient: $r = 0.280$, Spearman rank correlation: $\rho = 0.177$.}
\label{title-content}
\vspace{-12pt}
\end{figure}

\noindent 
In \textbf{variant detection test}, we managed to reach 58\% success rate for detecting variants (494 cases) and duplicates (1797) (see Figure \ref{title-content}). 23.1\% of failed cases (295 files) were of less than 20 records long, indicating a turbo tape \cite{garda2021} image with a loader\footnote{Turbo Systems on Atariki: \url{http://atariki.krap.pl/index.php/Systemy_turbo}}. Manual examination of the failed cases list revealed numerous cases of high Title Similarity (TS) values (cosine similarity) for the titles that would not be classified as the same by an operator, such as variants of River Raid or Boulder Dash, sequels -- like Joe Blade and Joe Blade II -- or multipart educational cycles. Some degree of confusion could be attributed to Montezuma's Revenge alone, which circulated in two versions, different in size and content, but under the same title\footnote{More on Montezuma's Revenge on FHKD: \url{https://fhkd.pl/2024/12/montezumas-revenge/}}. Some of the bootleg releases were compressed\footnote{More about the compression: \url{http://atariki.krap.pl/index.php/Cruncher_5.0}}, reducing file size and therefore changing the record pattern, while some were also XOR-encrypted\footnote{More on Turbo Copy 4 here: \url{https://github.com/seban-slt/tcx_tools} and here: \url{https://www.atari.org.pl/forum/viewtopic.php?id=12348&p=28}}, scrambling the content.

\begin{figure}[htbp!]
\centering
\includegraphics[width=\textwidth]{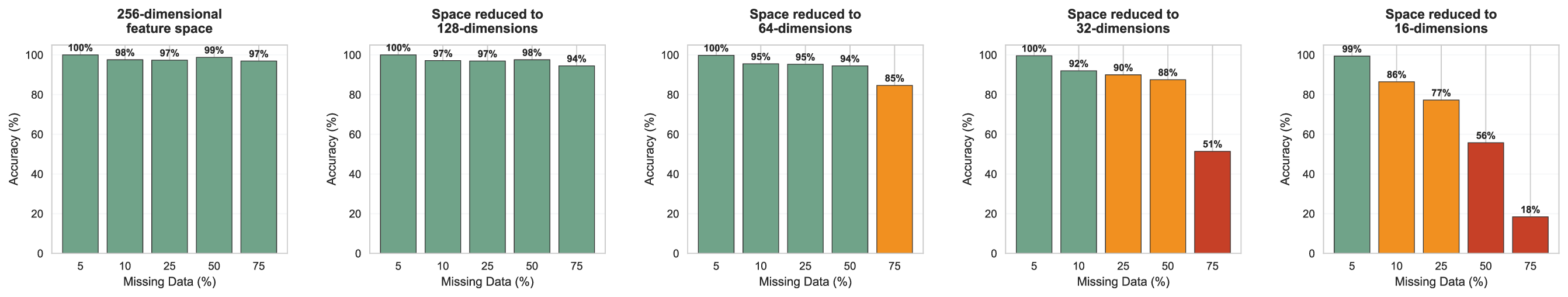}
\caption{Top-1 Accuracy degradation across vector representations. Top-1 accuracy measures the percentage of cases where the correct file (self) has the highest similarity score among all candidates. Green bars indicate >90\% accuracy, orange
70-90\%, and red <70\%. The accuracy decreases both with increased data loss
and reduced vector dimensionality, with 16-dimensional space showing critical failure at >50\% data loss.}
\label{top1-accuracy}
\end{figure}

\begin{figure}[htbp!]
\centering
\includegraphics[width=\textwidth]{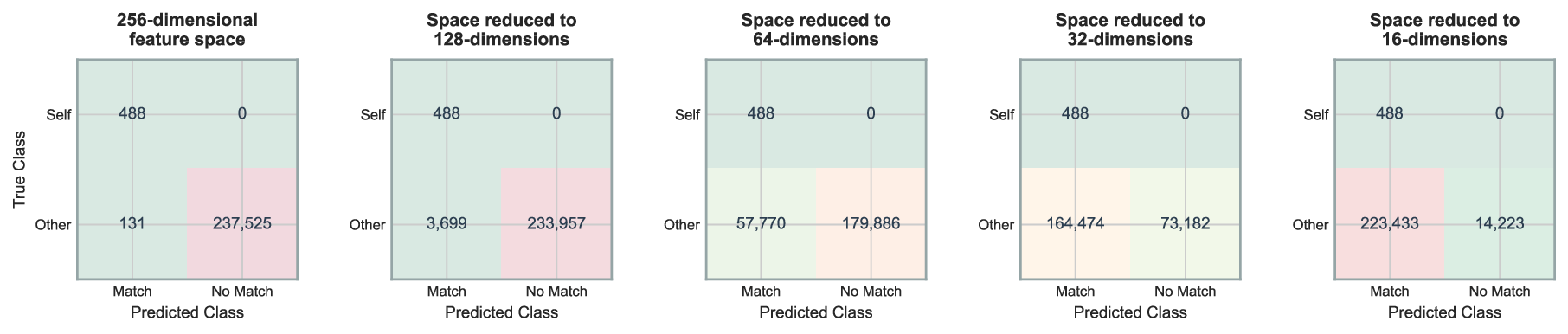}
\caption{Confusion matrices for 25\% data loss cases with match threshold at 0.7.}
\label{ok-confusion}
\vspace{-8pt}
\end{figure}

\begin{figure}[htbp!]
\centering
\vspace{-10pt}
\includegraphics[width=\textwidth]{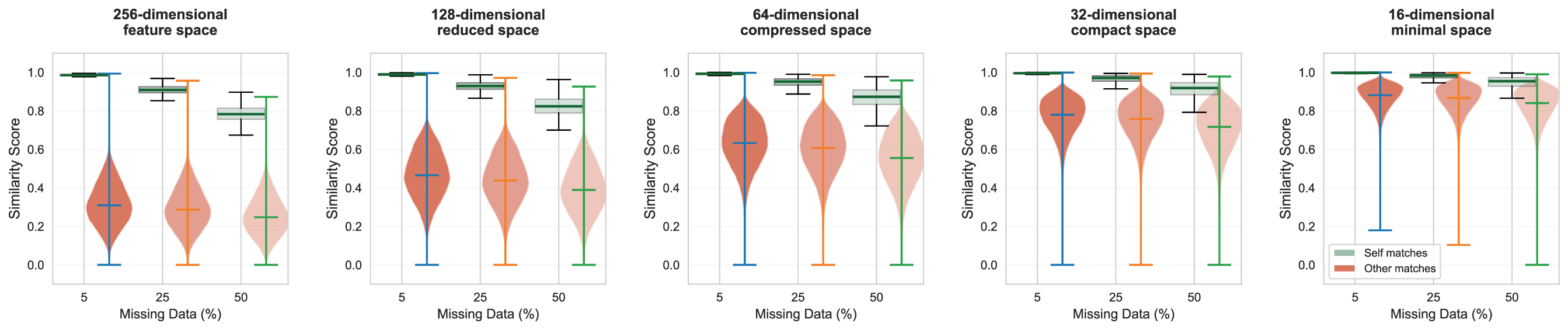}
\vspace{6pt}
\includegraphics[width=\textwidth]{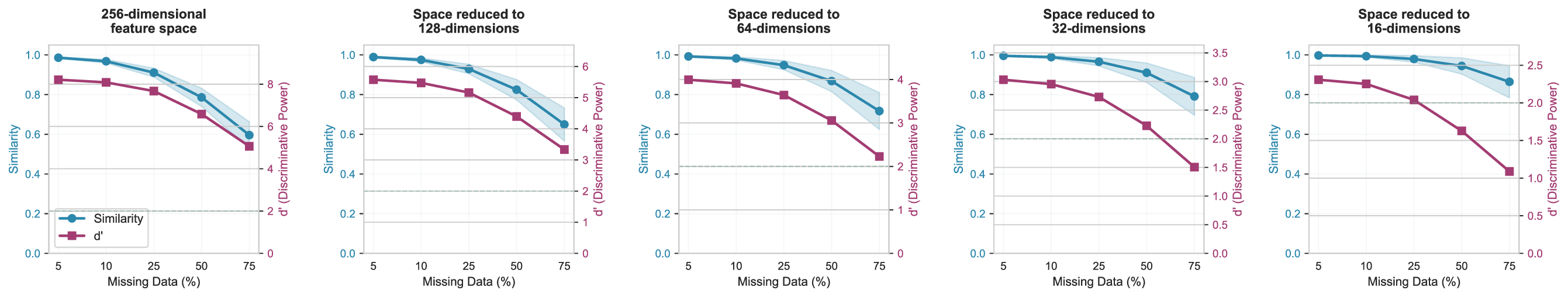}
\caption{\textbf{Top:} separation (d') between signal (self-matches) and noise (other-matches) distributions.
\textbf{Bottom:} mean similarity between degraded and original versions of the same file, with standard
deviation bands.}
\label{ok-distributions}
\end{figure}

\noindent 
In \textbf{Damaged copy recognition}, 256-dimensional model achieved 100\% accuracy for cases with minimal (5\%) data loss, with little degradation to 97\% accuracy for 75\% missing data cases (see Figure \ref{top1-accuracy}). We observed 100\% accuracy in recognizing self-samples at 25\% damage (488 cases) with no false negatives (see Figure \ref{ok-confusion}). The 256-dimensional model also has a low false positive rate of 0.05\% (131 cases), increasing with decreasing vector dimensionality (3699 cases for 128- and 57700 for 64-dimensional model). The similarity distributions maintain clear class separation, and the 256-dimensional model maintains a Discriminative Power above 5 even for cases in which 75\% of the data records were missing (see Figure \ref{ok-distributions}).

Results gathered during the first pass -- unmodified vectors, with peak 167 and 169 values present for the similarity calculations -- decayed more steeply as the data degradation progressed, with Discriminative Power (d') falling to 3.6 for 75\% damage (as compared to the second pass with d'=5). Accuracy was only slightly lower for 256-dimensions space, 10\% damage, at 97\% (98\% for second pass), and 95\% (as compared to 97\%) for 75\% damage. 

\section{Discussion}

\textbf{Performance.} The results confirm that the 256-dimensional Checksum Count Vector can be successfully used for the task of locating other copies of the recording in the repository (RQ1). Even in cases where data loss was significant, there was no practical impact on precision and recall (RQ2). Lowering the dimensionality below 256 results in no meaningful computational efficiency gain, yet negatively impacts accuracy (RQ3), indicating no curse of dimensionality. Although the best scores had been achieved in the damaged copy recognition experiment, where the high potential of the method is clearly visible, the duplicates detection ability is not to be undervalued. Even if only slightly more than half of the cases are correctly identified, this can still cut the workload in that area by the same proportion, freeing human experts to engage in more creative or rewarding tasks, such as repairing the most complex files manually or enriching the database with highly regarded personal stories \cite{10.1007/978-3-032-05802-7_28}.

\noindent
\textbf{Limitations.} While the vector similarity score alone cannot serve as a sufficient criterion for data merging or metadata integration, it is a robust mechanism for preliminary selection. The selected candidates must then be sequence-matched using strong checksum chains, followed by a final bit-by-bit comparison. The peaking values for all-zero records (see Figure \ref{dystrybucja-czeksum}) have 2-3\% negative influence on accuracy, hence it is advised, but not strictly required, to filter them out for best results. Finally, to have more representative numbers on data variants detection, higher quality metadata for the test set would be necessary - label confusion and turbo tapes in the mix influence the results negatively. For comparing turbo tapes, an extension of the method will be needed to adjust for the different record structure and sizes.

\noindent
\textbf{Integrity.} The \textit{principle of provenance} in archival science requires the preservation of information about the original origin of data \cite{Erez.2020}. Therefore, when using multiple sources for data reconstruction, care should also be taken to properly document both the sources and form of the data, as well as the detailed scope of their use. In this case, these could be the criteria based on which the data was matched, which fragments were added (with record-level accuracy), and the person responsible for the process, described as relationships in the CIDOC CRM model \cite{dorkhosh2021cidoc,10.1007/978-3-032-05802-7_28}. However, records duplication and incomplete or inaccurate metadata is a recurring theme \cite{grzeszczuk2023preserving,10.1007/978-3-032-05802-7_28}. Integration products should always be subject to \textit{human verification}, with appropriate documentation of the process. 

\noindent
\textbf{Future Work.} In parallel to this study, we work on automated file extraction, decoding and merging the products. This includes targeted approach for the problematic cases, with the use of iterative parameters selection and use of custom decoders, including machine learning based. A complete pipeline for processing standard FSK recordings could help convert individually hoarded tape piles into an integrated, cross-referenced, and museum-standard historical resource.

\section{Conclusions}

Effective and sustainable preservation of digital collections is not a task; it is a process. Long-term, time-consuming, fun, rewarding, necessary and best done as a group effort. But to encourage community participation, the tools used need to be as effective as possible. Our tests confirmed that Checksum Count Vectors can be successfully used to cross-reference and match resources and metadata between existing tape repositories. Such a robust mechanism for qualifying damaged sources for potential automatic repair can also be useful for creating new collections, which currently do not see the light of day due to a lack of time for individual volunteers to transform their work into a usable, accessible, and properly documented resource. In this way, our method supports scalable preservation, integrating metadata and improving the recovery of damaged files. This can help ease the process of bringing dispersed data together to form a more complete archive of early digital heritage.

\begin{credits}
\subsubsection{\ackname}The authors express their gratitude to the community of enthusiasts who tackle collection, maintenance, and preservation tasks, often without much systemic or institutional support. Leave no tape behind!

\subsubsection{\discintname}
The authors declare that they have no competing interests relevant to this research.
\end{credits}
%
%
%
\bibliographystyle{splncs04}
\bibliography{bibliography}

@misc{grzeszczuk2023preserving,
      title={{Preserving the Artifacts of the Early Digital Era: A Study of What, Why and How?}}, 
      author={Maciej Grzeszczuk and Kinga Skorupska},
      year={2023},
      eprint={2312.06570},
      archivePrefix={arXiv},
      primaryClass={cs.HC}
}

@inproceedings{parsons2004seti,
author = {Parsons, A and Werthimer, D and Anderson, David and Bowyer, Stuart and Cobb, Jeff and Demorest, P and Korpela, Eric and Lampton, Mike and Lebofsky, Matt},
year = {2004},
month = {01},
pages = {},
title = {{Searching for ET with help from four million volunteers: The SETI@home, SERENDIP, SEVENDIP, ASTROPULSE and SPOCK seti programs}},
volume = {4}
}

@article{moyle2010transcribe,
author = {Moyle, Martin and Tonra, Justin and Wallace, Valerie},
year = {2010},
month = {01},
pages = {},
title = {{Manuscript Transcription by Crowdsourcing: Transcribe Bentham}},
volume = {20},
journal = {Liber Quarterly},
doi = {10.18352/lq.7999}
}

@inbook{lekkas2014legalpirates,
author = {Lekkas, Theodore},
year = {2014},
month = {09},
pages = {73-103},
title = {{Legal Pirates Ltd: Home Computing Cultures in Early 1980s Greece}},
isbn = {978-1-4471-5492-1},
doi = {10.1007/978-1-4471-5493-8}
}

@ARTICLE{stachniak2015redclones,
  author={Stachniak, Zbigniew},
  journal={{IEEE Annals of the History of Computing}}, 
  title={{Red Clones: The Soviet Computer Hobby Movement of the 1980s}}, 
  year={2015},
  volume={37},
  number={1},
  pages={12-23},
  doi={10.1109/MAHC.2015.11}}

@inbook{wasiak2014playing,
author = {Wasiak, Patryk},
year = {2014},
month = {01},
pages = {129-150},
title = {{Playing and Copying: Social Practices of Home Computer Users in Poland during the 1980s}},
isbn = {978-1-4471-5492-1},
doi = {10.1007/978-1-4471-5493-8_6}
}

@article{aletras2013similarity,
author = {Aletras, Nikolaos and Stevenson, Mark and Clough, Paul},
title = {Computing similarity between items in a digital library of cultural heritage},
year = {2013},
issue_date = {December 2012},
publisher = {Association for Computing Machinery},
address = {New York, NY, USA},
volume = {5},
number = {4},
issn = {1556-4673},
url = {https://doi.org/10.1145/2399180.2399184},
doi = {10.1145/2399180.2399184},
journal = {J. Comput. Cult. Herit.},
month = {jan},
articleno = {16},
numpages = {19}
}

@InProceedings{10.1007/978-3-032-05802-7_28,
author="Grzeszczuk, Maciej
and Skorupska, Kinga
and W{\'o}jcik, Grzegorz Marcin",
editor="Biele, Cezary
and Kacprzyk, Janusz
and Kope{\'{c}}, Wies{\l}aw
and Mo{\.{z}}aryn, Jakub
and Owsi{\'{n}}ski, Jan W.
and Romanowski, Andrzej
and Sikorski, Marcin",
title={{Bridging the Digital Divide: Approach to Documenting Early Computing Artifacts Using Established Standards for Cross-Collection Knowledge Integration Ontology}},
booktitle="Digital Interaction and Machine Intelligence",
year="2026",
publisher="Springer Nature Switzerland",
address="Cham",
pages="273--280",
isbn="978-3-032-05802-7"
}

@journal-article{Erez.2020,
    author = "Erez, Sigal Arie and Blanke, Tobias and Bryant, Michael and Rodríguez, Kepa Joseba and Speck, Reto and Daelen, Veerle Vanden",
    doi = "10.1108/rmj-08-2019-0045",
    title = {{Record Linking in the EHRI Portal}},
    journal = "Records Management Journal",
    year = "2020"
}

@book{Crawford1982DeReAtari,
  title     = {{De Re Atari: A Guide to Effective Programming}},
  author    = {Crawford, Chris and Winner, Lane and Cox, Jim and Chen, Amy and Dunion, Jim and Pitta, Kathleen and Fraser, Bob and Makreas, Gus},
  publisher = {Atari Program Exchange},
  year      = {1982},
  address   = {Sunnyvale, CA},
  note      = {Original unbound three-hole punched pages},
  url       = {https://archive.org/details/ataribooks-de-re-atari}
}

@article{six2022panako,
author = {Six, Joren},
year = {2022},
month = {10},
pages = {},
title = {Panako: a scalable audio search system},
volume = {7},
journal = {Journal of Open Source Software},
doi = {10.21105/joss.04554}
}

@inproceedings{DBLP:conf/ismir/Wang03,
  author       = {Avery Wang},
  title        = {{An Industrial Strength Audio Search Algorithm}},
  booktitle    = {{ISMIR} 2003, 4th International Conference on Music Information Retrieval,
                  Baltimore, Maryland, USA, October 27-30, 2003, Proceedings},
  year         = {2003},
  bibsource    = {dblp computer science bibliography, https://dblp.org}
}

@inproceedings{rahutomo2012cosine,
author = {Rahutomo, Faisal and Kitasuka, Teruaki and Aritsugi, Masayoshi},
year = {2012},
month = {10},
title = {{Semantic Cosine Similarity}}
}

@article{weber2023digital,
  title={{Introduction to the Special Issue on Digital Natural and Cultural Heritage: Opportunities and Challenges}},
  author={Weber, Andreas and Heerlien, Maarten and Gassó Miracle, Eulàlia and Wolstencroft, Katherine},
  journal={ACM Journal on Computing and Cultural Heritage},
  volume={16},
  number={1},
  pages={1e},
  year={2023},
  doi={10.1145/3597459}
}

@book{swalwell2021,
  author    = {Melanie Swalwell},
  title     = {Homebrew Gaming and the Beginnings of Vernacular Digitality},
  year      = {2021},
  publisher = {MIT Press},
  address   = {Cambridge, MA},
  isbn      = {9780262044776},
  url       = {https://mitpress.mit.edu/9780262044776},
}

@unknown{afforce_2022,
author = {Skorupska, Kinga and Nielek, Radoslaw and Kopeć, Wiesław},
year = {2022},
month = {07},
pages = {},
title = {{AFFORCE: Actionable Framework for Designing Crowdsourcing Experiences for Older Adults}}
}

@inproceedings{zooniverse_2023,
author = {Jeong, Eunmi (Ellie) and Jackson, Corey and Dowthwaite, Liz and Ahmad, Tallal and Trouille, Laura},
title = {{Assessing the Value Orientations of Contributors to Virtual Citizen Science Projects}},
year = {2023},
isbn = {9798400707582},
publisher = {Association for Computing Machinery},
address = {New York, NY, USA},
url = {https://doi.org/10.1145/3593743.3593782},
doi = {10.1145/3593743.3593782},
booktitle = {Proceedings of the 11th International Conference on Communities and Technologies},
pages = {191–202},
numpages = {12},
location = {Lahti, Finland},
series = {C\&T '23}
}

@article{chng2023photogrammetry,
author = {Ch'ng, Eugene},
year = {2023},
month = {10},
pages = {},
title = {Engaging institutions in crowdsourcing close-range photogrammetry models of urban cultural heritage},
journal = {Journal of Cultural Heritage Management and Sustainable Development},
doi = {10.1108/JCHMSD-07-2022-0107}
}

@article{wirtz2025savingukraine,
author = {Wirtz, Gudrun},
year = {2025},
month = {09},
pages = {251-253},
title = {{Saving Ukrainian Cultural Heritage Online (SUCHO)}},
volume = {54},
journal = {Preservation, Digital Technology \& Culture},
doi = {10.1515/pdtc-2025-0049}
}

@article{dauterive2025save,
author = {Dauterive, Jessica and Mitchell, Mary},
year = {2025},
month = {09},
pages = {31-49},
title = {{Save What You Can: Tending Katrina’s Community Archive}},
volume = {31},
journal = {Southern Cultures},
doi = {10.1353/scu.2025.a968661}
}

@article{capaccioni2025borndigital,
author = {Capaccioni, Andrea},
year = {2025},
month = {09},
pages = {1-11},
title = {Digitized and born-digital cultural heritage: implications for digital humanities},
volume = {16},
journal = {JLIS.it},
doi = {10.36253/jlis.it-657}
}

@article{vries2016commodore,
author = {Vries, Denise and Harrington, Craig},
year = {2016},
month = {12},
pages = {},
title = {{Recovery of heritage software stored on magnetic tape for Commodore microcomputers}},
volume = {11},
journal = {International Journal of Digital Curation},
doi = {10.2218/ijdc.v11i2.386}
}

@article{pereda2025onlineheritage,
author = {Pereda, Javier and Willcox, Pip and Candela, Gustavo and Sanchez, Alexander and Murrieta-Flores, Patricia},
year = {2025},
month = {03},
pages = {39-69},
title = {Online cultural heritage as a social machine: a socio-technical approach to digital infrastructure and ecosystems},
volume = {7},
journal = {International Journal of Digital Humanities},
doi = {10.1007/s42803-025-00097-6}
}

@article{Lichnerowicz_Grzeszczuk_Skorupsk_2024, 
   title={{Wyzwania i możliwości w zakresie ochrony niematerialnego dziedzictwa kulturowego: wnioski z Demosceny}}, 
   volume={63}, 
   url={https://apcz.umk.pl/LSE/article/view/55083}, 
   DOI={10.12775/LSE.2024.63.12}, 
   journal={Łódzkie Studia Etnograficzne}, 
   author={Lichnerowicz, Andrzej and Grzeszczuk, Maciej and Skorupska, Kinga}, 
   year={2024}, 
   month={wrz.}, 
   pages={217–234} 
}

@unknown{sadia2025unstructured,
author = {Sadia, Mushtari and Chowdhury, Amrita and Chen, Ang},
year = {2025},
month = {09},
pages = {},
title = {{A Case for Computing on Unstructured Data}},
doi = {10.48550/arXiv.2509.14601}
}

@article{dorkhosh2021cidoc,
author = {Dorkhosh, Maliheh and Fattahi, Rahmatollah and Arastoopoor, Sholeh},
year = {2021},
month = {02},
pages = {},
title = {{Extensions of CIDOC-CRM: Responses to the Need for Knowledge Organization in Subject Domains}},
doi = {10.30484/NASTINFO.2020.2497.1942}
}

@inbook{garda2021,
author = {Garda, Maria and Grabarczyk, Paweł},
year = {2021},
month = {05},
pages = {37-55},
title = {{“The Last Cassette” and the Local Chronology of 8-Bit Video Games in Poland}},
isbn = {978-3-030-66421-3},
doi = {10.1007/978-3-030-66422-0_3}
}

\end{document}